%% file: main.tex
\documentclass[letterpaper, 10 pt, conference]{ieeeconf}  

\IEEEoverridecommandlockouts                              
\usepackage[utf8]{inputenc}
\usepackage[T1]{fontenc}
\usepackage{xcolor}
\usepackage{amsmath}
\usepackage{soul}
\usepackage{booktabs} 
\usepackage{graphicx}
\usepackage{amssymb}
\usepackage{comment}
\usepackage{hyperref}

\title{\LARGE \bf
Explosive Legged Robotic Hopping: Energy Accumulation and Power Amplification via Pneumatic Augmentation} 

\author{Yifei Chen, Arturo Gamboa-Gonzalez, Michael Wehner, and Xiaobin Xiong  
\thanks{*This work was supported in part by the University of Wisconsin-Madison Office of the Vice Chancellor for Research and Graduate Education funded by the Wisconsin Alumni Research Foundation. The authors are with the department of Mechanical Engineering, the University of Wisconsin, Madison. Corresponding author: Xiaobin Xiong {\tt\small xiaobin.xiong@wisc.edu}. The experiment video can be seen here: 
\href{https://youtu.be/JObkOIaiOqE}{https://youtu.be/JObkOIaiOqE}}
}

\begin{document}
\newcommand{\bb}[1]{\mathbb{#1}}
\newcommand{\todoArturo}[1]{{\textbf{Arturo, do this:\color{red} #1}}}
\newcommand{\tomove}[1]{{\color{blue} #1}}
\newcommand{\todelete}[1]{{\color{green} #1}}

\newcommand{\smallsection}[1]{{\vspace{1mm} \noindent{\textbf{#1}:}}}

\newcommand{\smallitsection}[1]{{\vspace{1mm} \noindent{\underline{#1}:}}}

\maketitle
\thispagestyle{empty}
\pagestyle{empty}

\begin{abstract}
We present a novel \textit{pneumatic augmentation} to traditional electric motor-actuated legged robot to increase intermittent power density to perform infrequent explosive hopping behaviors. The pneumatic system is composed of a pneumatic pump, a tank, and a pneumatic actuator. The tank is charged up by the pump during regular hopping motion that is created by the electric motors. At any time after reaching a desired air pressure in the tank, a solenoid valve is utilized to rapidly release the air pressure to the pneumatic actuator (piston) which is used in conjunction with the electric motors to perform explosive hopping, increasing maximum hopping height for one or subsequent cycles. We show that, on a custom-designed one-legged hopping robot, \textit{without} any additional power source and \textit{with} this novel pneumatic augmentation system, their associated system identification and optimal control, the robot is able to realize highly explosive hopping with power amplification per cycle by a factor of approximately 5.4 times the power of electric motor actuation alone. 

\end{abstract}


\section{INTRODUCTION}




Legged robots, with the ability to make and break ground contact, select contact locations, and modulate gait are able to traverse a broader range of terrains than wheeled or tracked robots. They have shown great potential in search and rescue \cite{bellicoso2018advances}, inspection \cite{gehring2021anymal}, and exploration in unstructured environments \cite{agha2021nebula}. Although legged robots can have different numbers of legs, joints, and morphologies, their locomotion behaviors are often cyclic, requiring negative work in each cycle of the periodic motion. Taking hopping robots \cite{LeggedRobotThatBalance} as examples, before accelerating upwards to lift off the ground, the robots must decelerate to a complete stop in the vertical direction in the descending phase by performing negative work on its center of mass.

Significant research activities in the literature have been focused on using elastic elements in the system to prevent using actuators to perform negative work, increasing energy efficiency, and reducing the required actuator power in the design. The elastic elements are typically in the form of air springs \cite{LeggedRobotThatBalance} or mechanical springs \cite{grizzle2009mabel, brown1998bow, haldane2017repetitive, hubicki2016atrias, wang2023terrestrial}. The negative work in the descending phase can be stored as spring potential energy and returned to the system in the ascending phase of the cyclic motion. The springs are installed either in series \cite{pratt1995series, yang2023small, EPA_hopper} or parallel \cite{Disney3D, 9636108} with the main actuators on the robots; despite this difference, the elastic energy must be stored and released immediately within one cycle, which prevents the \textit{periodic energy accumulation for power amplification} to achieve subsequent emergent explosive maneuvers.

\begin{figure}[t]
    \centering
    \includegraphics[width=\linewidth]{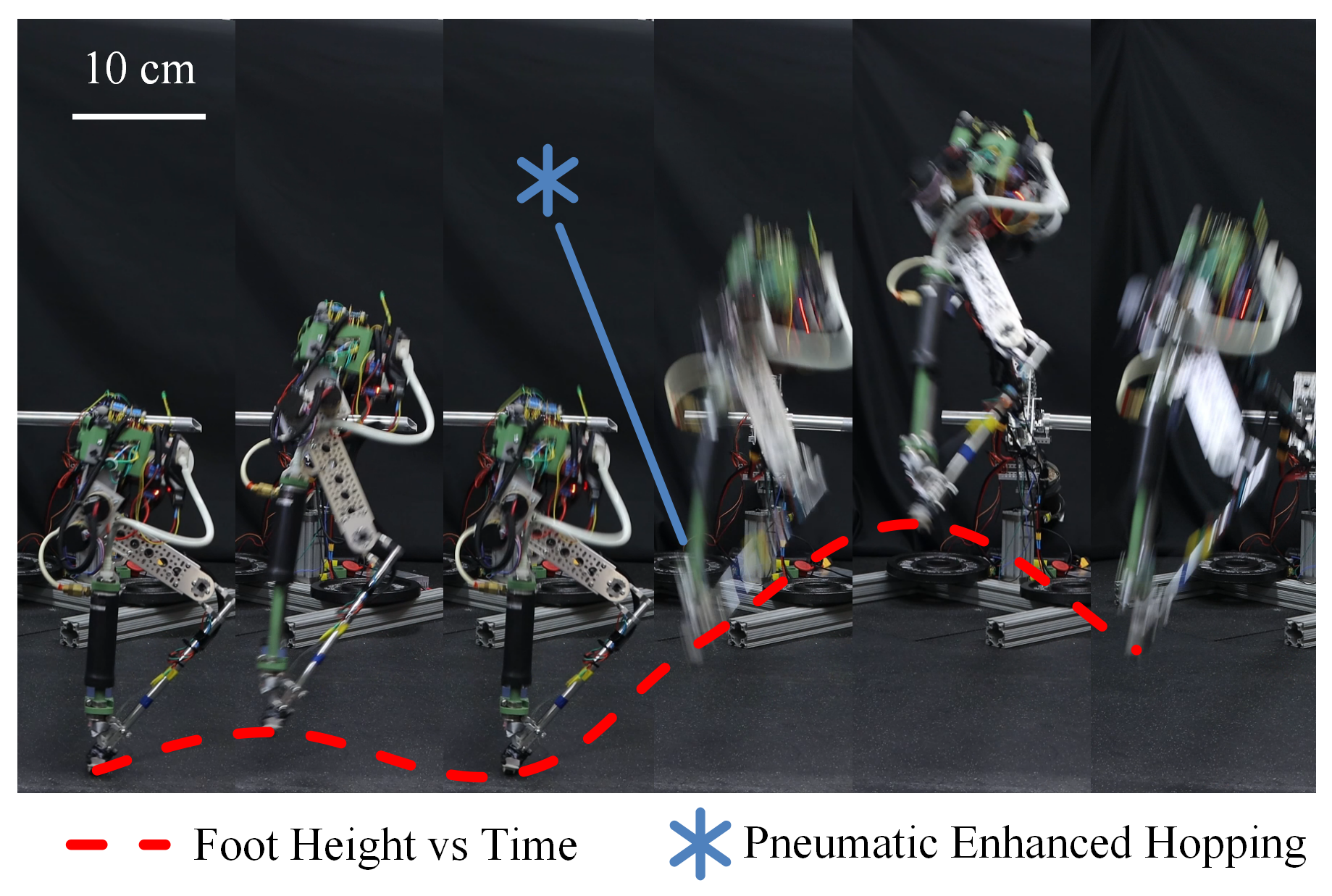}
    \caption{The explosive jumping (right) that is "warmed up" by the regular periodic jumping behaviors (left) of the single-legged hopping robot.}
    \label{fig:Robot design structure}
\end{figure}



To address this problem, we propose a design and control framework that augments legged robots with a pneumatic system to accumulate energy during the periods of negative work for storage over multiple cycles. With this system, \textit{a pump} converts this kinetic energy to potential energy in the form of compressed air pumped to \textit{a tank}. As an integrated part of the system, no special pumping phase is necessary, as the pump performs its energy conversion during normal locomotion activities. Each cycle pumps additional air, increasing the potential energy in the tank. When the tank has reached a desired pressure (sufficient potential energy), the air can be released, rapidly displacing \textit{a pneumatic actuator} (converted back to kinetic energy) which is utilized in conjunction with the original actuator to amplify power output for explosive locomotion behaviors.

We realize this framework on a single-legged hopping robot that is originally actuated only by electrical motors. We custom-design the pneumatic augmentation system based on the associated physical parameters of the robot. The control of hopping with the pneumatic system is rigorously realized via model-based techniques. With careful system identifications of the pneumatic system, its dynamics can be combined with the canonical rigid-body dynamics model \cite{lynch2017modern} of the robot. We then use a direct collocation method \cite{hereid20163d, xiong2018bipedal, hubicki2016tractable} to optimize the trajectories for pneumatically-enhanced (and non-enhanced) hopping behaviors based on a reduced order model. Task space position and force controls are deployed on the robot to realize these hopping behaviors. 

For this specific robot, we show that with the custom-design pneumatic system, the robot increases its hopping height from $\sim4.3$ cm to $\sim23.4$ cm, that is equivalent to amplifying power per cycle by 5.4 times. Moreover, the stored energy enables the robot to perform challenging maneuvers, e.g. jumping onto a high platform, without additional power sources. The results from the physical system validate our proposed approach and indicate a promising design and control paradigm for power amplifications on highly dynamic-legged robots.

\input{design}

\input{modeling}

\input{control}

\input{experiment}

\section{Conclusions and Future Work}
We present a pneumatic augmentation framework for traditional legged robots that are originally actuated by electric motors. A custom-designed one-legged robot is used for validating the framework. The pneumatic system utilizes a pump to convert negative work done during locomotion cycles into energy that is stored in a tank as air pressure. With accumulated air pressure in the tank, a pneumatic actuator is then utilized to perform high-power-output jumping behaviors that cannot be done by the original electric motors. 

In the future, we are interested in further optimizing and extending the design and control framework. Presumably, the tank can be integrated into the actuator for maximizing the efficiency and power density. The pump and actuator with carefully controlled valves can be treated as springs that store and release energy over multiple cycles with flexibility in modulating energy flow in each cycle, i.e. dissipating or injecting a controllable amount of energy into the system. This combined with appropriate control methodologies can potentially enable legged robots to extend their operation spectrum with low motorization requirements.

\addtolength{\textheight}{-0cm}   



 
\bibliographystyle{ieeetr}
\bibliography{References}

\end{document}

%% file: design.tex
\section{System Design}

\par 
We start by describing the architectural design of our pneumatically augmented hopping robot. With a goal of using the proposed pneumatic augmentation to increase the power density of traditional legged robots, we build our initial augmentation-free robot from the open-sourced robot Hoppy \cite{9636108} with modest modifications. The pneumatic augmentation is then designed to append onto the robot with both mechanical, electrical, and software integrations. 

\subsection{Barebone Robot Design}
\begin{figure}[t]
    \centering
    \includegraphics[width=\linewidth]{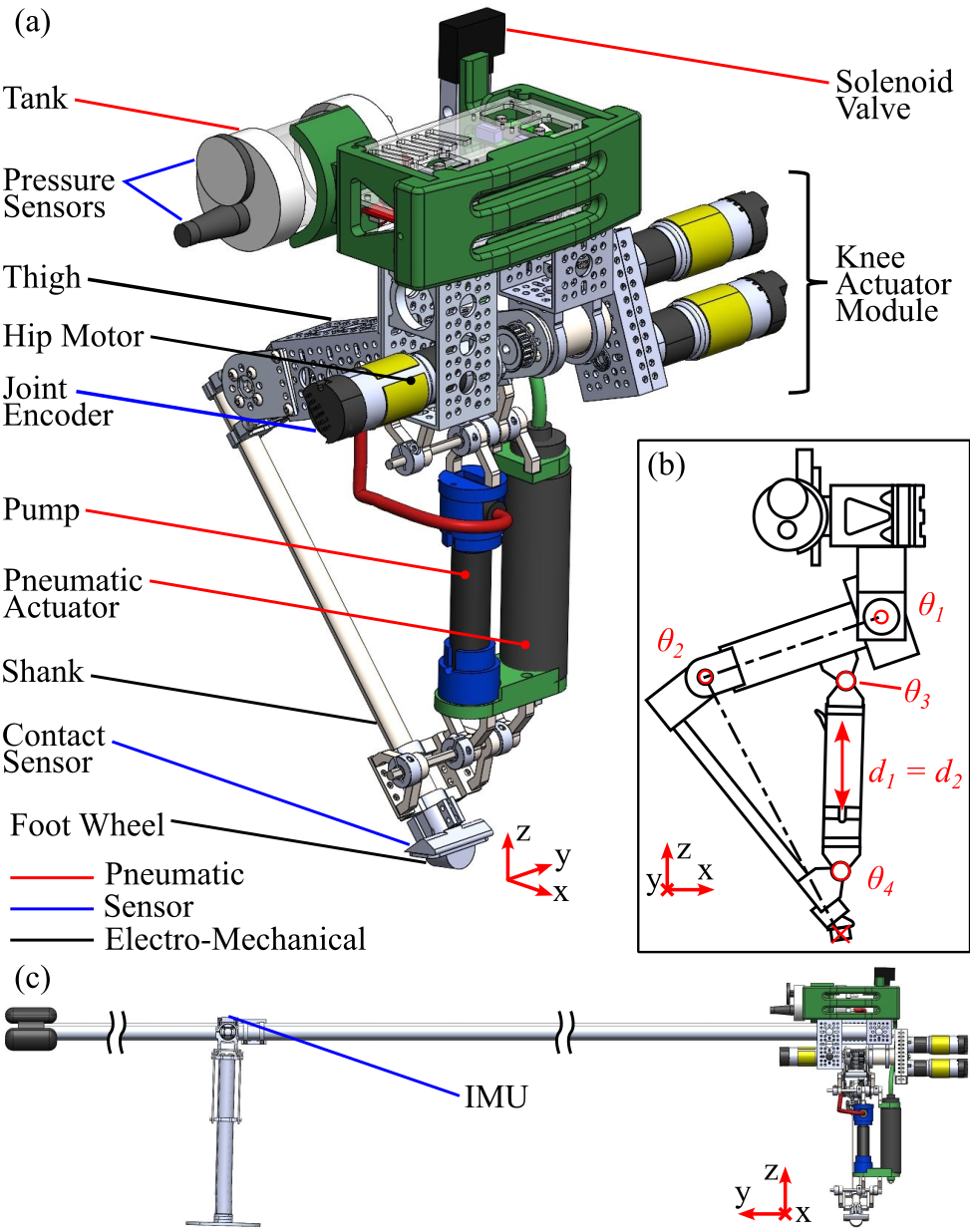}
    \caption{ System. (a) Key components. (b) Linkage model and the degrees of the freedom (DoFs) of the robot. (c) Planarizing boom. The weight of the robot itself is 3.7 kg, 700g of which is the pneumatic system weight. 
    }
    \label{fig:Robot design structure}
\end{figure}

\noindent{\underline{\text{Mechanical Components:}}} The robot is a single legged hopper with two linkages and two joints, as shown in Fig. \ref{fig:Robot design structure} (b). Similar to \cite{9636108}, brushed DC motor modules and off-the-shelf modular parts from Gobilda are selected to actuate the joints and serve as the linkages, respectively; the robot's motion is also planarized using a boom of similarly modular parts as illustrated in Fig. \ref{fig:Robot design structure} (c). As we would like to enable the robot to jump without significant counter-weight, we carefully select the available DC motor modules (max. speed 223 rpm, max. torque 3.728 Nm). More importantly, we use two DC motor modules together (connected by gears) to actuate the knee since the knee joint requires higher joint torques during dynamic jumping. Additionally, the knee motors are positioned at the upper hip to reduce the leg inertia; the motor rotation is then transmitted to the knee joint via a belt drive with a 1.5:1 reduction ratio. We removed the parallel springs in \cite{9636108} in our design to leave space for the direct use of our pneumatic augmentation. Last, we added a wheel under the foot to allow the foot sliding along the boom direction.


\vspace{1mm}

\noindent{\underline{\text{Electrical Components:}}} To implement real-time control, we use a TI LaunchPad microcontroller (LAUNCHXL-F28379D, 200MHz dual C28xCPUs) that is programmed in Simulink (C2000 package) Fig. \ref{fig: Digram_Ele_pneu} (a). This board provides sufficient computation capacity and low-level interfaces to sensors and actuators, which realizes a target control frequency of 1 kHz on the robot. 
The brushed DC motors are controlled through PWM via VNH5019 Motor Drivers (12 A, 5.5-24 V). The motors has integrated rotatory encoders (751.8 pulses per revolution), which is directly read using the TI board. A force sensing resistor (FRS, FlexiForce A201, 11 kg) is installed on the foot to detect contact transitions between ground and aerial phases during hopping. An IMU (SparkFun ICM-20948 9DoF IMU) is installed on the top part of the boom to measure the rotational velocity of the boom that can be used to calculate the local linear velocity of the robot. We use a joystick controller (VOYEE wired PC controller) connected to an Arduino UNO via a USB-C host shield (SparkFun USB-C Host Shield). This setup allows for commanding locomotion task parameters in realtime. The Arduino UNO processes the signals from the IMU and the joystick controller, and transmits data to the TI board through serial communication.

\subsection{Pneumatic Augmentation}
Our system eschews springs, which store energy momentarily, in favor of pneumatics, to store energy for any desired duration. While the components in this system were selected for this robot, the underlying concept is applicable to most legged robots and myriad systems with non-constant contact between two surfaces.

\vspace{1mm}
   
\begin{figure}[t]
    \centering
    \includegraphics[width=\linewidth]{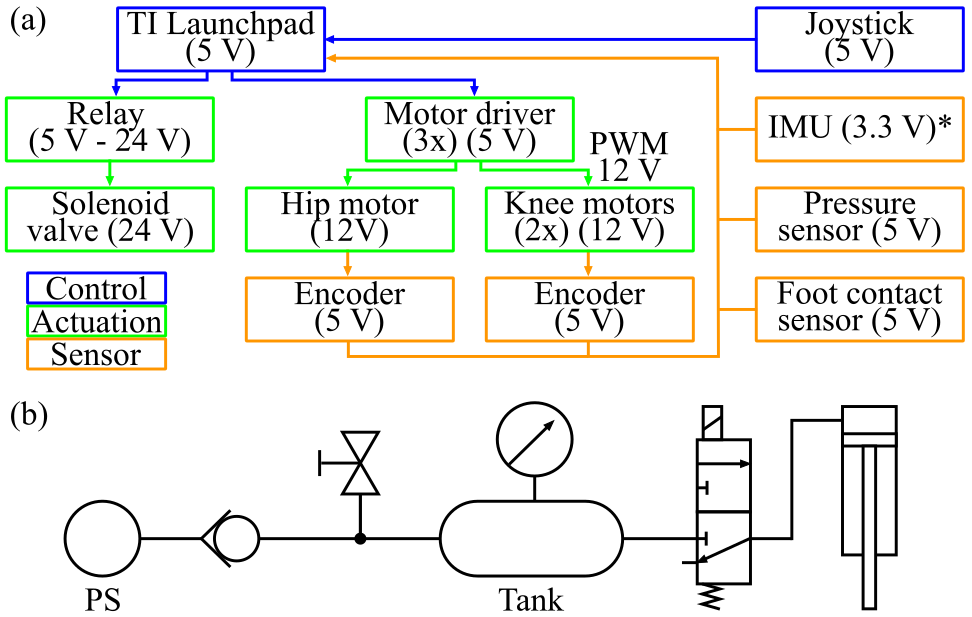}
    \caption{ (a) System Schematic. *IMU remotely mounted on boom and processed through a secondary microcontroller. (b) Schematic of pneumatic system. Components form left to right: Pressure Source (pump), check valve (one way valve), manual ball valve (safety release), tank with pressure sensor, three-way solenoid valve, pneumatic actuator.}
    \label{fig: Digram_Ele_pneu}
\end{figure}

\noindent{\textbf{System:}} 
The system is shown in Fig. \ref{fig:Robot design structure} (a), and schematically in Fig. \ref{fig: Digram_Ele_pneu} (b). A two-stage piston pump (Giyo GM043) pumps air past a check-valve (which prevents pressurized air from flowing back to the pump) into a tank, where pressure is monitored by two sensors (one analog, one digital). After pressure reaches a pre-set target pressure, a solenoid valve (Norgren V60) can be opened to quickly release air to a pneumatic actuator (Airpot 2KS325). The pneumatic actuator is mounted alongside the pump, thus stroke for the two are set to be identical with a maximum displacement of 105 mm. 

\noindent{\textbf{Pump-Actuator:}} 
A piston pump displaces a piston inside of a cylinder to compress air. Our two-stage (telescoping) piston pump includes two concentric pistons (stages) to increase stroke in a small form factor (Fig. \ref{fig:Pneumatic Pump Model}). 
During pumping, the first stage (working diameter, 14 mm) is actuated, raising the internal air pressure (Fig. \ref{fig:Pneumatic Pump Model}, $100-l_{1}$). After a brief transition, the second stage (working diameter, 17 mm) is actuated, further raising the pressure of internal air following the Compressing curve (Fig. \ref{fig:Pneumatic Pump Model} $l_{2}$). When pump pressure exceeds tank pressure, the check valve opens and air is pumped into the tank (labeled Pumping in Fig. \ref{fig:Pneumatic Pump Model}). Prior to this Pumping, all work has been done to compress the gas, and no energy has been added to the tank. With a goal of maximizing force during Pneumatic Actuation, we selected a large diameter (32.5 mm) pneumatic actuator ($force = pressure \cdot area$). 

\vspace{1mm}
\noindent{\textbf{Tank Design and System Balancing:}} 
As released air displaces the pneumatic actuator, system volume (tank + actuator) increases and overall pressure decreases, reducing the force available to do work. Thus, target tank volume was at least twice the max actuator volume. The final tank (mass, 205 g) has a volume 2.3x actuator volume. Small diameter (1.5 mm id) pump-to-tank tubing was selected to minimize the volume of air to be compressed before the check valve is opened. A large solenoid valve and large diameter tubing (6 mm id) were chosen between the tank and the pneumatic actuator to maximize flow rate and actuation speed. In all cases, the shortest practical tubing lengths were chosen. A pump with small piston diameters (14, 17 mm) was selected generated high pressures with moderate applied force ($pressure = force \div area$). The two-stage design increased overall stroke length, increasing pumping rate.

%% file: modeling.tex
\section{Modeling}
\label{sec:modeling}
In this section, we first present the dynamics model of the pneumatic augmentation, and then the hybrid dynamics of the hopping behaviors of the robot. The leg contraction and extension  for realizing hopping originally is controlled by the DC motors; with the pneumatic  system, the leg contraction is impeded by the resistance of the pump and friction of the actuator, and the leg extension is impeded by the friction of the pump and enhanced by the pneumatic actuator. Therefore, to methodically control hopping, we first identify the dynamics of the pneumatic augmentation.

\subsection{System ID of Pneumatic Augmentation}

\subsubsection{Pumping Resistive Force Model}

The pump has different resistive forces in extension and compression. Let $d$ denote the pump length. During extension $\dot{d} > 0$, the pressure in the pump $P_\text{pump}$ is constant at atmosphere, so there is only trivial friction force. During compression $\dot{d} < 0$, the resistive force changes based on the length of $d$, and more importantly, on whether is check-valve is being pushed \textit{open} ($P_\text{pump} > P_\text{tank}$) or remains \textit{closed} ($P_\text{pump} < P_\text{tank}$), where $P_\text{tank}$ denotes the pressure in the tank. We present a theoretical model to calculate the pressure change in the pump to calculate the resistive force during compression, and then a date-driven model based on actual measurement of the resistive forces on an experimental apparatus.  

\smallsection{Theoretical Model}
Before the check-valve is pushed open during compression, the pressure $P_\text{pump}$ inside the pump follows the Boyle's Law, indicating the product of $P_\text{pump}$ and the pump chamber volume $V_\text{pump}$ is constant: 
\begin{equation} \label{Boyles_Law}
    P_\text{pump}(d) V_\text{pump}(d) = P_0 V_0
\end{equation}
where $P_0$ is the atmospheric pressure, and $V_0$ is the initial volume of the pump. In compression, $V_\text{pump}$ decreases and thus $P_\text{pump}$ increases. The resistive force of the pump during compression with check-valve remaining closed thus is:
\begin{equation} \label{Pump_Pressure_1}
    F^\text{closed}_\text{pump}(d)= P_\text{pump}(d) A_\text{pump} = \frac{ P_0V_0}{d  },
\end{equation}
where $A_\text{pump}$ is the cross-sectional areal of the pump. Here we assume $V_\text{pump} = A_\text{pump} d$.

As the pump is pushed to a critical compression distance, denoted by \(d_C = \frac{P_0 V_0}{P_\text{tank} A_\text{pump}}\), such that $P_\text{pump} \geq P_\text{tank}$, the check-valve is pushed open and connects the pump to the tank, which increases the total volume. The force required to continually compress of the pump thus becomes:
\begin{equation} \label{Pump_Pressure_2}
  F^\text{open}_\text{pump}(d)=A_\text{pump}\frac{P_\text{tank} (V_\text{tank}+A_\text{pump}d_C) }{V_\text{tank}+A_\text{pump}d}
\end{equation}
where $V_\text{tank}$ is the volume of the tank. 

\begin{figure}[t]
    \centering
    \includegraphics[width=\linewidth]{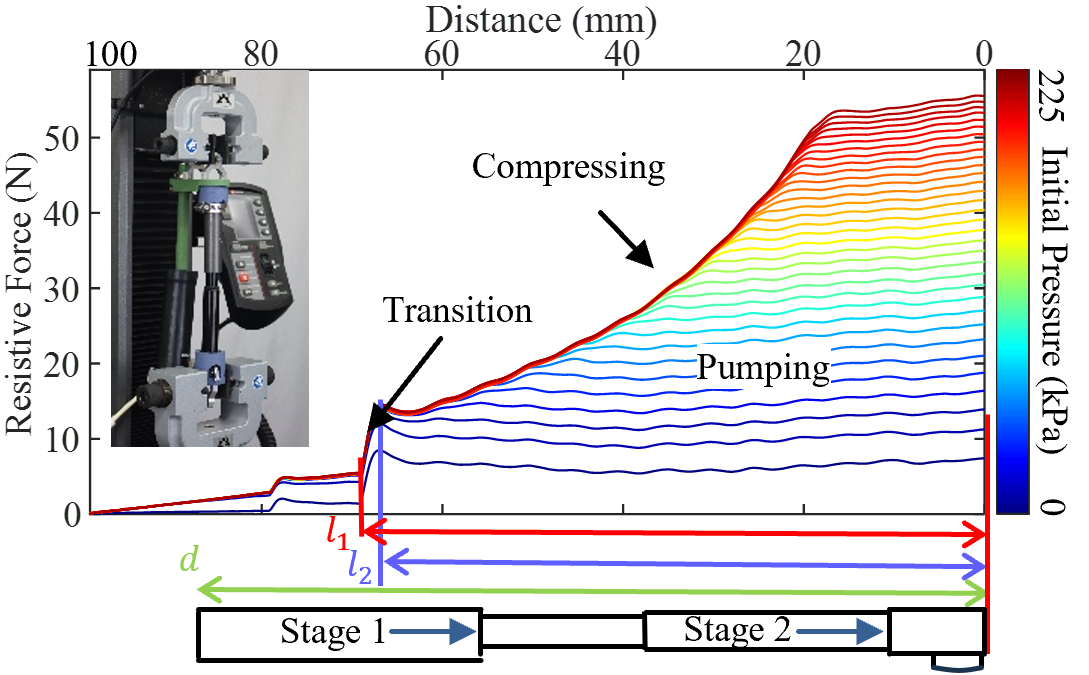}
    \caption{The resistive forces of the pneumatic pump are plotted vs. the compression distance. The pump moves between Stage 1 and Stage 2, with a short transition distance between the two stages. The pump resistive force follows a single curve (Stage 1) until the check-valve is triggered, at which point the force curve deviates as it is adding pressure to the tank (Stage 2). Additionally, the pump resistive force in Stage 2 depends on the initial tank pressure. The embedded photo shows the experiment setup of the pump on a tensile tester.}
    \label{fig:Pneumatic Pump Model}
\end{figure}

\smallsection{Data-Driven Model} 
To characterize the Force-Displacement response of the two-stage pump at various tank pressures, we install the pump with the tank on a tensile tester (Instron, 5943), and perform pumping at constant speed until tank pressure reaches 225 kPa (Fig. \ref{fig:Pneumatic Pump Model}).

The data indicates the resistive forces before the check-valve opens are more complex during to the two stages. The Stage 1 (\(d<l_1\)) raises pressure less and thus requires less force to compress than Stage 2. Additionally, the transition interval (\(l_1\le d<l_2\)) sees a sharp but relatively linear increase in forces. Finally, Stage 2 (\(l_2\le d<l\)) follows the curve labeled as Compressing in Fig. \ref{fig:Pneumatic Pump Model} which we estimated as a 2nd order polynomial. Therefore, the force before the check-valve opens is approximated as: 
\begin{equation} \label{Pump_Pressure_3}
    \tilde{F}^\text{closed}_\text{pump}(d)=\begin{cases}
        m_1 d + b_1 & d > l_1 \\
        m_2 d + b_2 & l_1 \ge d > l_2 \\
        c_1d^2+c_2d+c_3 & d \le l_2
    \end{cases}
\end{equation}
where 
the coefficients in the polynomials are fitted from the data. After the check-valve is triggered $d\ge d_C$ and the pump begins to add pressure to the tank, the force required to continue compress the pump closely approximates the theoretical curve from \eqref{Pump_Pressure_2} until it reaches the end of the stroke, yet we approximate it as a second order polynomial for accuracy. Taking into consideration all of the cases results in the pump's force dynamics being:
\begin{equation} \label{Pump_Force_Lumped}
    F_\text{pump}(d)=\begin{cases}
        \tilde{F}^\text{closed}_\text{pump}(d) & d<d_C, \dot{d}<0 \\
        \tilde{F}_\text{pump}^\text{closed}(d_C)(c_4d^2+c_5d+c_6) & d\ge d_C, \dot{d} <  0  \nonumber 
    \end{cases}
\end{equation}


\subsubsection{Actuator Dynamics Model}
As the pump is building pressure in the tank, the pneumatic actuator acts as a passive prismatic joint with minimal friction. Once the tank pressure reaches a desired level, the solenoid valve can be triggered allowing airflow from the tank to the pneumatic actuator which extends the leg. The exerted actuation force is modeled by combining two parts: 1. a \textit{quasis-static model} that only depends on the actuator pushed length $d$, and 2. a \textit{transient dynamics model} that considers the force dynamics in the duration of the air pressure being equalized after the solenoid valve is triggered. 

\begin{figure}[t]
    \centering
    \includegraphics[width=\linewidth]{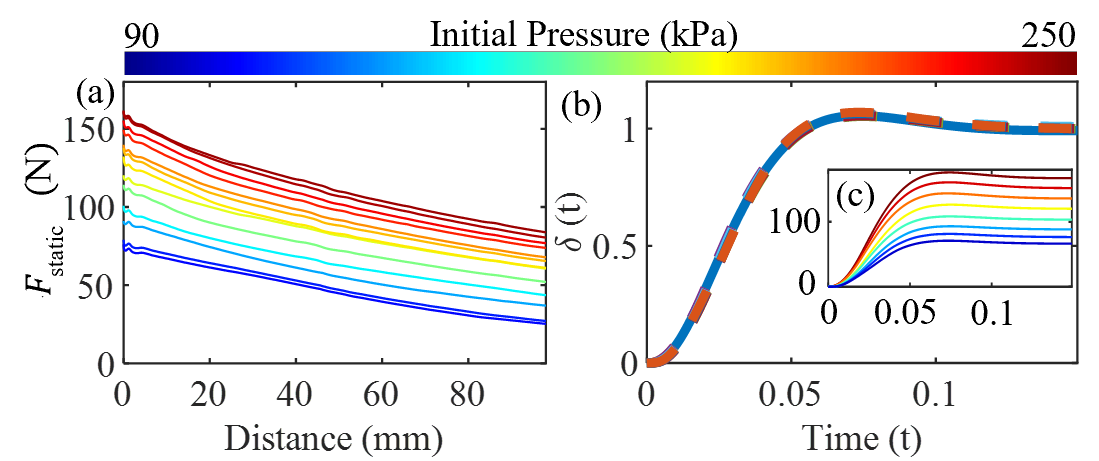}
    \caption{(a) The forces that the pneumatic actuator exert as a function of leg distance for different initial tank pressures. (b) The transient force response of the pneumatic actuator after the solenoid valve is triggered. }
  
    \label{fig:Pneumatic Acutator}
\end{figure}

\smallitsection{Quasis-static Model} The pneumatic pushing force first decreases with the increase of the pushed length $d$, which behaves similarly to the theoretical pumping model \eqref{Pump_Pressure_2}, because pushing is quasis-statically equivalent to pumping. The pushing force is thus calculated by:
\begin{equation} \label{Actuator_Force_1}
F_\text{static}(d)=A_\text{pa}\frac{P_\text{tank} V_\text{tank}}{V_\text{tank}+A_\text{pa}d}
\end{equation}
where $A_\text{pa}$ is the cross-section area of the pneumatic actuator. The same tensile tester is used to directly measure the force on the actuator when it is being extended. As shown in Fig. \ref{fig:Pneumatic Acutator} (a), the resulting force exerted by the pneumatic actuator decreases as a function of actuation distance, which matches with \eqref{Actuator_Force_1}.

\smallitsection{Transient Dynamics Model} When the solenoid valve is triggered, it takes a non-negligible amount of time for the air to reach to the actuator. For highly dynamic jumping motion, it is desirable to consider these transient force dynamics. We use the mechanical tensile tester to measure this dynamic force at the moment of triggering the solenoid valve. Fig. \ref{fig:Pneumatic Acutator} (b) shows, as that regardless of the initial tank pressure, the transient response of the force can be approximated the same 2nd order ordinary differential equation (ODE) by treating triggering the valve as a step input. The dynamics of the actuation force is thus approximated by: 
\begin{equation}\label{Actuator_Transient}
k_1 \ddot{\delta} + k_2 \dot{\delta} + k_3 \delta = s_v(t)
\end{equation}
where $\delta$ is the force state that is normalized by $F_\text{static}(d=0)$, $k_1$, $k_2$, and $k_3$ are constants identified using the Matlab System ID. toolbox on the experimental data, and $s_v(t)$ denotes the step input from the solenoid valve.

\smallitsection{Combined} To better approximate the pneumatic actuator force dynamics in leg extension, both the transient response and the kinematics based force mode are combined:

\begin{equation}\label{Pneum_Actuator_Force_total}
    F_\text{pa}(t,d)=F_\text{static}(d)\delta(t) 
\end{equation}

\subsubsection{Lumped Pneumatic Force} Because the pump and the pneumatic actuator are installed in parallel and can be modeled as a single prismatic joint, the total force is:
\begin{equation} \label{eq:pneu_lum}
    F^{\sum}_{\text{pneu.}} 
    =
    \begin{cases}
        F_\text{pump}(d) &   \dot{d}<0 \\
                             0&   \dot{d}>0, s_v = 0. \\
   F_\text{pa}(t,d) &   \dot{d}>0, s_v = 1
     \end{cases}
\end{equation}

\subsection{Hybrid Dynamics of Hopping}
 
The identified pump-actuator models are directly integrated into the actuator dynamics within the canonical rigid-body dynamics model of the robot. We model the pump and the pneumatic actuator together as a prismatic joint that have different actuation force models depending whether the robot is using the pump to store energy or the actuator to do explosive hopping. The robot thus has a closed-loop kinematic chain, which is modelled similarly to that of robot Cassie \cite{xiong20223}. This robot still has two self degrees of freedom (DOFs), where the leg length is actuated both by the knee motors and the pneumatic actuator. For hopping behaviors, the dynamics is hybrid that is composed of two phases: an aerial phase and a stance phase. The transition from  aerial phase to  stance phase is a discrete impact map, whereas the transition from  stance to  aerial phase is smooth. 


\vspace{1mm}
\noindent{\underline{\textit{Continuous Dynamics}:}} Both the aerial phase and stance phase have continuous dynamics, the Euler-Lagrangian equations of which can be compactly represented by:
\begin{equation} \label{eq:continuousDynamics}
    M(q)\ddot{q} + C(q,\dot{q})+ G(q) = B_m \tau_m + B_\text{p} F^{\sum}_\text{pneu.} + J_\text{h}^T F_\text{h}, 
\end{equation}
 where  $M(q)$ denotes the mass matrix, $C(q, \dot{q})$ is Coriolis and centrifugal forces, $G(q)$ represents the gravitational torque vector, $\tau_m$ is the motor torque, $F^{\sum}_\text{pneu.}$ denotes the lumped pneumatic force in \eqref{eq:pneu_lum}, $B_m$ and $B_\text{p}$ represent the associated actuation matrices, $F_\text{h}$ is the holonomic constraint forces, and $J_\text{h}$ is the Jacobian of the holonomic constraints.

The holonomic constraints include the closed loop chain constraints in the leg. In the stance phase, the foot ground contact is assumed to be non-slipping in the forward direction, which introduce additional holonomic constraints. Compactly,
    $J_\text{h}\ddot{q} + \dot J_\text{h} \dot{q} = 0$, 
which is combined with \eqref{eq:continuousDynamics} to describe the continuous dynamics for the stance and aerial phases with differences on the set of the holonomic constraints. 

\vspace{1mm}
\noindent{\underline{\textit{Discrete Transitions}:}} We assume the impact between the foot and the ground at touch-down is purely plastic. The post-impact state satisfies the holonomic constraints, and the impulse of the impact creates a change of momentum.
\begin{equation}
    \begin{bmatrix}
        \dot{q}^+ \\ 
        F_{\Delta}
    \end{bmatrix}
     = 
     \begin{bmatrix}
         M(q) & -J_\text{h}^T \\
         J_\text{h} & 0_{2 \times 2} \\
     \end{bmatrix}^{-1}
     \begin{bmatrix}
         M(q)\dot{q}^{-} \\
         0_{2 \times 1}
     \end{bmatrix},
\end{equation}
where $F_{\Delta}$ is the impulse force from the ground, and $ ^+$ and$^-$ represent the states at post- and pre- impact, respectively. 




%% file: control.tex
\section{Control Synthesis}
\label{sec:control}
 
With the dynamics being identified, we now present our method of motion planning and feedback control for realizing dynamic hopping behaviors on the pneumatically augmented legged robot. 
We define two kinds of hopping tasks: a regular periodic hopping that realizes energy storage, and an explosive hopping behavior that maximize hopping height. The trajectories of motion are optimized with different cost functions and constraint variations, and they are realized with the same low-level control methods.

\subsection{Control Strategy}

The control of hopping is decoupled into two sub-tasks: \textit{vertical control} to maintain a periodic hopping height, and \textit{horizontal control} to stabilize a forward velocity. The horizontal control is realization in the aerial phase by choosing the target leg angle: 
$ 
    q^\text{des}_\text{leg} = k_p (v_k - v^d) + k_d (v_k - v_{k-1})
$ 
where $v_k$ is the horizontal velocity of the center of mass (COM) of the robot at the apex event at $k$ step, and $v^d$ is the desired forward hopping velocity, and $k_p$ and $k_d$ are the feedback gains. For vertical jumping behaviors, $v^d =0$. The desired leg angle is selected as one output. An additional output in the aerial phase is on the leg length. The desired leg length $L^\text{des}$ is set to be the maximum leg length so that at touch down, the robot can utilize the full stroke length of the pump to maximize pumping. The desired output trajectories are then tracked via a task-space PD controller.

The vertical height control is realized on the ground phase, where it has to retract and then extend the leg to accelerate itself to lift off with certain vertical velocity for reaching to a target apex height. To best utilize the pneumatic augmentation while actuating the electrical motors, we apply trajectory optimization techniques to plan for optimal jumping behaviors. The optimized actuation force trajectories are then realized via a task-space force controller.

\begin{figure}[b]
    \centering
    \includegraphics[width=0.9\linewidth]{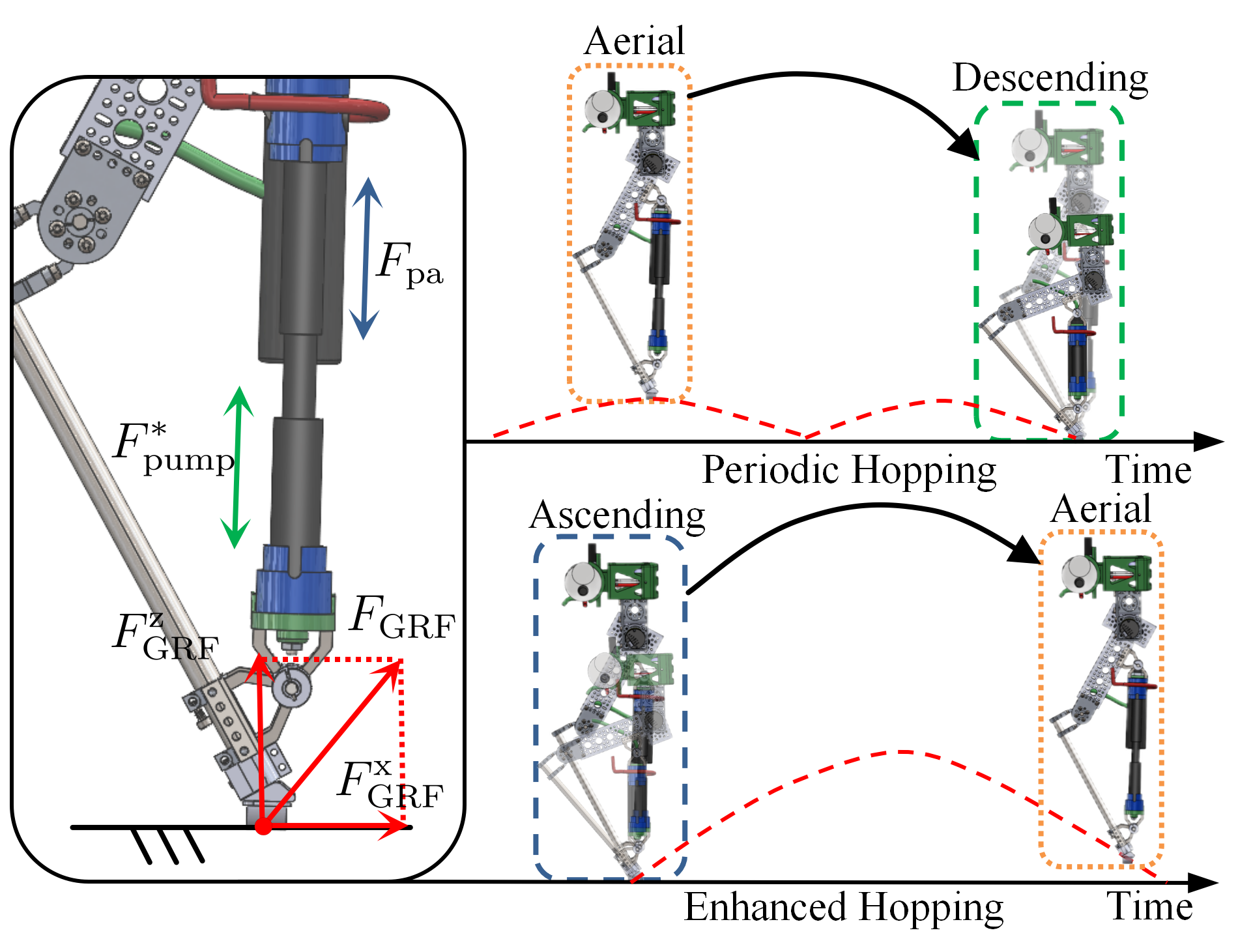}
    \caption{ Illustration of the hybrid dynamic model and control methods. In the stance phase, force control ensures precise ground interaction, while task space PD control governs the aerial phase for agile maneuvering.}
    \label{fig:HybirdDynamics}
\end{figure}


\subsection{Trajectory Optimization}

As the robot is top-heavy with small leg inertia, we utilize a point mass model, that is placing the COM of the robot at the hip, to plan the actuation force of the motors and pneumatic actuator for realizing optimal jumping maneuvers via trajectory optimization using direct collocation. Here, we list the dynamics, constraints, and cost functions of the problems. Readers can refer to \cite{hubicki2016tractable, xiong2018bipedal} for the details of the implementation for similar problems. 

\smallsection{Simplified Dynamics} Considering vertical hopping behaviors, the COM on the ground phase is actuated under the actuating force from the electrical motors and forces of the pneumatic system. The dynamics are:
\begin{equation}
    \tilde{m} \ddot{L}=
    F^{\sum }_\text{pneu.} + F_m - mg
    \label{Forces_Sum}
\end{equation}
where $L$ is the vertical leg length between the COM and the foot, $g$ is the constant gravitational term, and $F_m$ is the vertical force contributed from the electrical motors. $\tilde{m}$ is the projected mass of the whole system to the hip position, including the robot and the boom, which is calculated based on equivalent kinetic energy.  Since $F^{\sum }_\text{pneu.}$ has discontinuity at $\dot{d} = 0$ where the leg length transitions from compression to extension, we divide the ground phase into \textit{descending} and \textit{ascending} phases. We use trapezoidal collocation to approximate the dynamics on the ground phase. The aerial dynamics is purely ballistic, so solutions of its trajectories can be calculated in closed-form. The transitions between the phases are assumed to be smooth.

\smallsection{General Constraints} First, $L$ should satisfy the range of motion of the robot: $L_\text{min} \leq  L \leq L_\text{max}$. The ground reaction force should be non-negative for all time, and approaches 0 at lift-off. Then we assume an initial apex height $H_\text{apex}$ where the robot  starts to fall before the ground phase.  Additionally, it is expected that the leg length will be fully extended before the ground phase in order to fully compress the pump to inject more pressure to the tank. Thus, the initial state in the ascending phase should satisfy:
   $  L_\text{descending}^i = L_\text{max} ,
    (\dot{L}_\text{ascending}^i)^2 = 2g (H_\text{apex} - L_\text{max}) \nonumber$,
where $^i$ denotes the initial state. The vertical velocity is 0 at the transition between ascending and descending. Additionally, it is desired that the height reaches to its lowest point for maximum compression:
   $  L_\text{descending}^f  = L_\text{ascending}^i= L_\text{min}, 
    \dot{L}_\text{descending}^f = \dot{L}_\text{ascending}^i = 0
 \nonumber$, where $^f$ denotes the final state.

\smallsection{Periodic Hopping} 
The final state of ascending should realize the same apex height for periodic hopping:
    $ (\dot{L}_\text{ascending}^f)^2 =  2g(H_\text{apex} -   L_\text{ascending}^f). \nonumber$ 
The cost function is defined to minimize the actuator's control effort:
    $J_\text{cost} = \smallint{ F^2_m dt}.$
Solving this optimization problem yeilds the optimized actuation force that is used to provide optimal hopping motion while charging up the tank starting at a known starting air pressure $P_\text{tank}$. 
Ideally, the optimization would be solved for each hopping cycle to realize consistent periodic motion, because as tank pressure changes, the pump force profile changes.
In practice, we instead optimize the motion once and use the following to calculate actuation force for the subsequent hopping motion:
$ 
    F^{k}_m = F^{*}_m + F^{*}_\text{pump} -  F^{k}_\text{pump}
$
where $^*$ denotes the optimized solution and $k$ denotes the $k$ th periodic hopping cycle. Since the pneumatic actuator is not used during periodic hopping, only the pump force is acted on the prismatic joint, and it can be calculated based on the starting tank pressure of that hopping cycle.

\smallsection{Explosive Hopping} The solenoid valve is set to be open in the ascending phase to utilize the pneumatic actuator for explosive hopping. The cost function is designed to generate maximum apex height while keeping a smooth force profile:
$J_\text{cost} = \textstyle c \int{ F^2_m dt} - \ln{(\frac{{\dot{L}^{f^2}_\text{ascending}}}{2g}+L_\text{ascending}^f)}$,
 where $c$ is the cost coefficient.

\subsection{Realtime Feedback Control}
The desired trajectories are realized via task-space controllers to calculate the desired motor torques, which are realized by the motor controller via sending PWMs. 

\subsubsection{Task-Space Position Control in Aerial Phase}
\par
During the aerial phase, the objective is to control the leg length and leg angle to the desired values. The desired trajectories are planned as described in section \ref{sec:control} A. We use a task-space PD controller to calculate the required joint torques:
    $\tau_\text{aerial} = J_y^{-1} (K_p(y - y^\text{des}) + K_d (\dot y  - \dot y^\text{des}))$,
where $y = [q_\text{leg}, L]^T$ is the output that includes the leg angle and leg length, and $K_p$ and $K_d$ are the PD control gains.

\subsubsection{Task-space Force Control in Stance Phase}
During the stance phase, we synthesize a kinematics based force controller to realize the optimized motor force profile $F^{*}_m(t)$, which is combined with the forces on the pneumatic systems to realize the vertical ground reaction force $F^{z}_\text{GRF}$ for jumping. The pneumatic actuator is controlled by turning the solenoid valve at the optimal timing. Additionally, when horizontal motion is required, we use a Bézier curve to generate a desired force profile $F^{x}_\text{GRF}$(t) in the horizontal direction to assist with the foot placement controller. Here, the peak of this force profile is linearly proportional to the desired forward hopping velocity.  
Assuming the movement of the lightweight leg has a negligible effect on the joint space dynamics, we use the Jacobian mapping to translate required motor actuation forces from the task-space to the required motor torques in the joint-space: $ 
    \tau_\text{stance} = J_f^{T}   \begin{bmatrix}
        F^{x*}_\text{GRF}, 
        F^{*}_m 
    \end{bmatrix}^T, 
$ 
where $J_f$ represents the Jacobian of the foot position w.r.t. the hip. Combining the stance and aerial phase together yields
$ 
    \tau _m= \alpha \tau_\text{stance} + (1 - \alpha) \tau_\text{aerial},
$
where $\alpha$ is either 0 to 1 based on the contact measurement.

\subsubsection{Motor Controller}
To convert the desired motor torques $\tau_m$ into the required input voltages ($\text{V}_\text{PWM}$) for the motors, we apply motor dynamics equations:
$ 
    \text{V}_\text{PWM} = \frac{R_{\omega}}{k_T N} \tau + k_e N\dot{\theta},
$ 
where $R_{\omega}$ denotes the coil resistance, $k_T$ is the motor torque constant, and $N$ represents the gearbox speed reduction ratio, $k_e$ is the motor electrical constant ($\text{V}\cdot \text{s/rad}$), and $ \dot{\theta}$  is the joint velocity. For our experiments, we take a conservative approach by assuming $k_e \approx 0 $.

%% file: experiment.tex
\section{Results}
 

\subsection{Pneumatically Enhanced Hopping}

\begin{figure}[t]
    \centering
    \includegraphics[width=\linewidth]{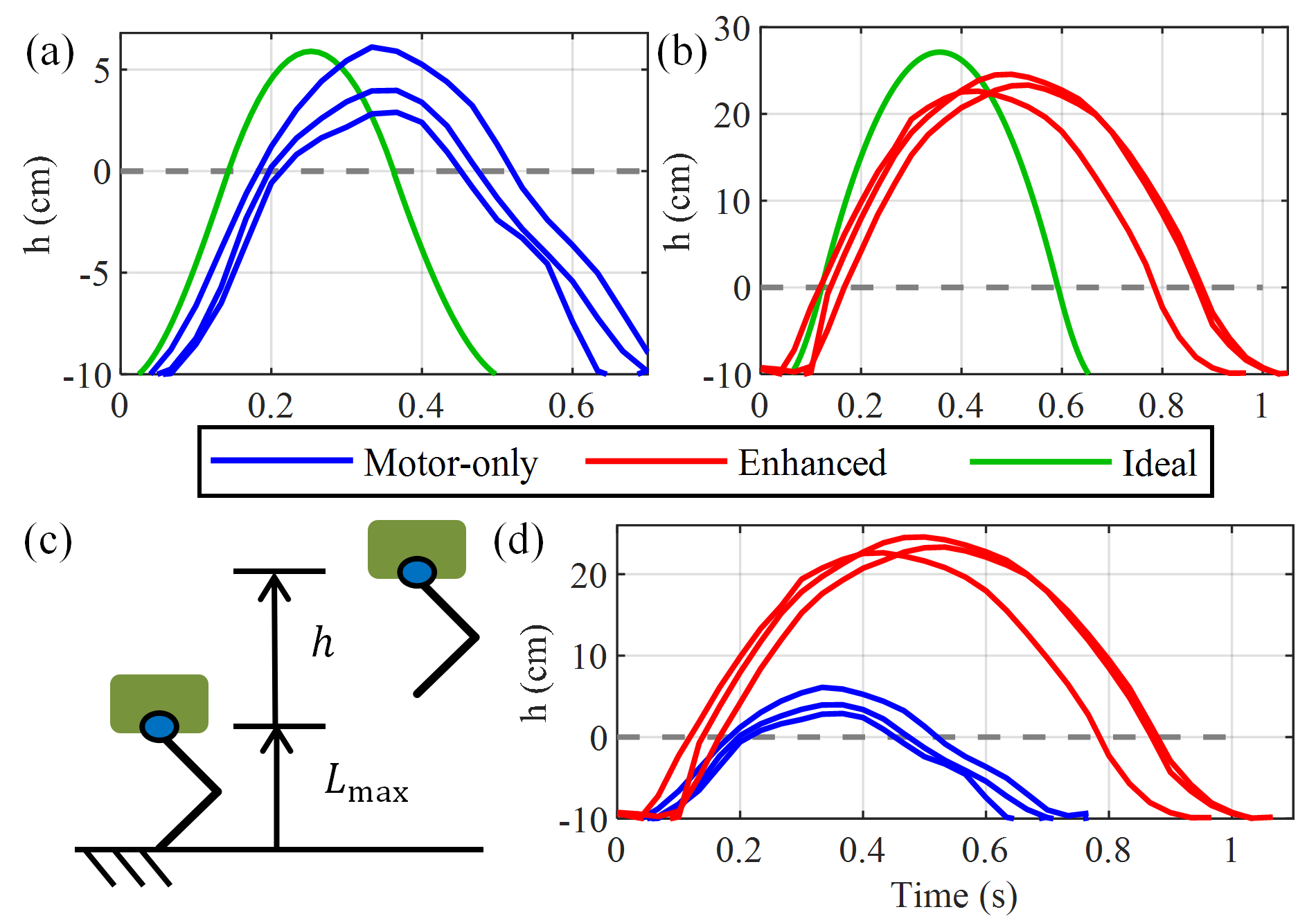}
\caption{Comparison of the trajectories of the hip during (a) the maximum height hopping by only using motors, and (b) pneumatically enhanced hops released at 303.37 kPa. (c) The height is defined as the hip joint height measured from the leg being fully extended. (d) Comparison of the trajectories between the pneumatically enhanced-free and enhanced hopping.}
    \label{fig:TrajectoriesFig}
\end{figure}

We first realize the control framework for hopping with maximum apex height only using the electric motors; the manual ball valve is turned open so that the the pump is not connected to the tank. The robot is set to have a perceived static weight of 2.2 kg. Fig. \ref{fig:TrajectoriesFig} (a) shows the realized hopping trajectories on the robot compared with the optimized one from our model. The vertical apex height reaches average 4.3 cm (max at 6.1 cm, min at 2.9 cm) which matches with the expected height to a good extent; we deem that the low-quality DC motors are the cause of the performance variance. The aerial phase on the hardware is shorted in duration because the projected gravitation on the actual robot is smaller as the robot jumps up on the boom. 

\begin{figure}[t]
    \centering
    \includegraphics[width=\linewidth]{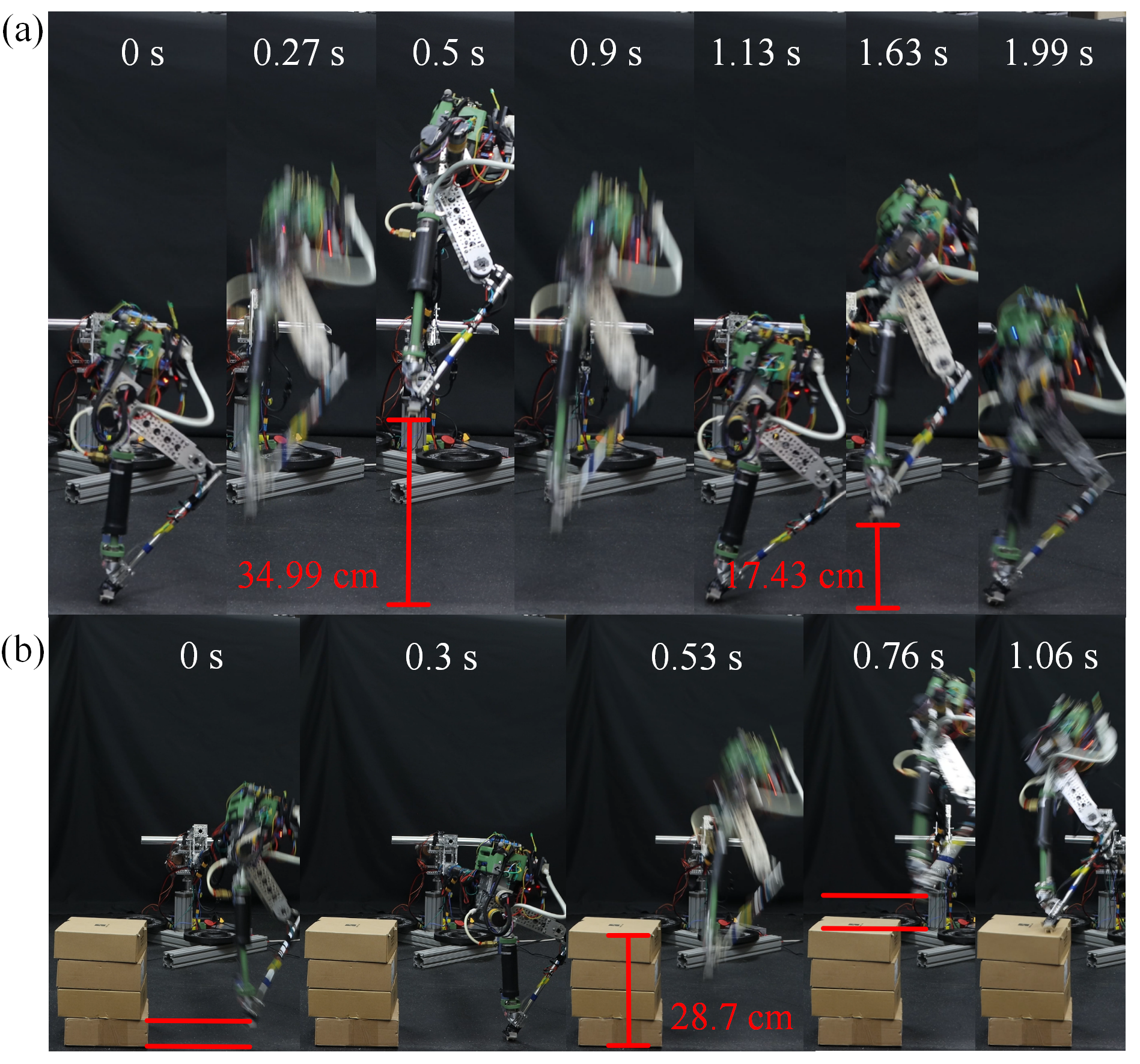} 
    \caption{ (a) Consecutively enhanced hopping: release all the stored pressures with two consecutive enhanced hops. (b) Jumping onto a high platform where an pneumatically enhanced-free jump fails ($m=$1.9 kg).}
    \label{fig:extra_behavior}
\end{figure}

Then we realize the framework with using the pneumatic appendage on the robot to storage energy in periodic hopping cycles and then optimize explosive behaviors. Fig. \ref{fig:TrajectoriesFig} (b, c) shows the realized pneumatically enhanced hopping trajectories. The optimized trajectory that combines pneumatic and electric motor actuation precisely predicts the apex height, and
the robot is able to jump up to average 23.4 cm (max at 24.6 cm, min at 22.6 cm) consistently with the same tank pressure, which indicates a power amplification per hopping cycle with a factor of $\sim5.4$ x. Moreover, the stored energy during regular hopping can be used for realizing other high-powered behaviors such as consecutively enhanced hopping and jumping onto a platform, as shown in Fig. \ref{fig:extra_behavior}.




\subsection{Performance Analysis}   
\begin{figure}[t]
    \centering
    \includegraphics[width=\linewidth]{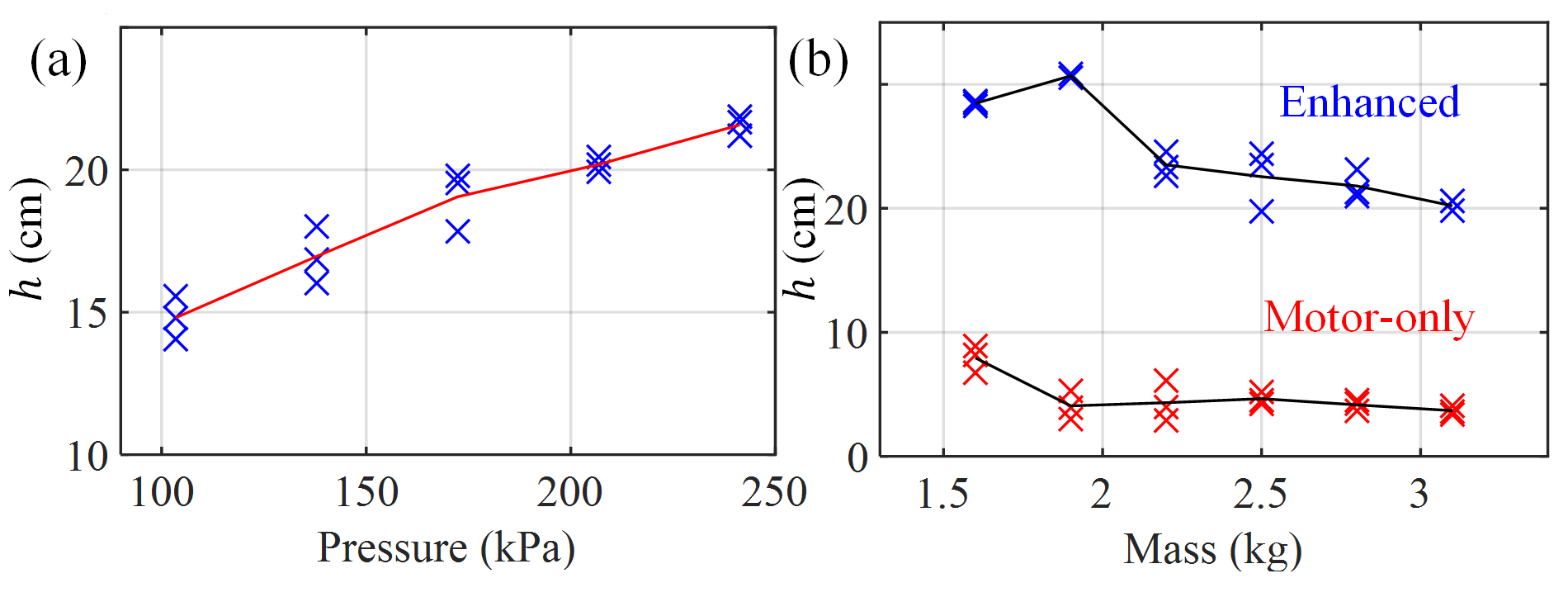}
    \caption{Apex height of the hip of enhanced hopping behaviors with (a) varying release tank pressures with robot weight at 2.2 kg and (b) varying robot weights of where the robot performs the enhanced hopping when the pressure in the tank plateaued, i.e., reaches to the maximum value. The red marks in (b) are the maximum height realized by the electric motors alone without the pneumatic system being installed on the robot, which produces much lower hopping height. }
    \label{fig:HeightVaryingParams}
\end{figure}

The air pressure value in the tank represents the amount of stored energy from periodic hopping behaviors. Fig. \ref{fig:HeightVaryingParams} (a) shows that the apex height of the enhanced hopping increases linearly with the stored air pressure in the tank, as expected. However, the maximum tank pressure is limited by the maximum pumping force during regular hopping cycles, which is inherently limited by the original electric motors and the weight of the robot. We thus explore the weight effect of the robot on the realizable maximum apex height by the enhanced hopping; the results are shown in Fig. \ref{fig:HeightVaryingParams} (b). As the robot is mounted on the boom with counter-weights, we adjust the location of the counter-weights which changes the perceived static weight of the robot. The experiments show that, for the same electric actuation with the pneumatic augmentation, the maximum achievable aerial height increases with the increases of the robot weight, peaks at 1.9 kg, and then decreases as the robots get heavier. This suggests that the value of design optimization at the system level, balancing energy storage, efficiency, and power density.


\subsection{Energy Analysis}
\begin{figure}[t]
    \centering
    \includegraphics[width=\linewidth]{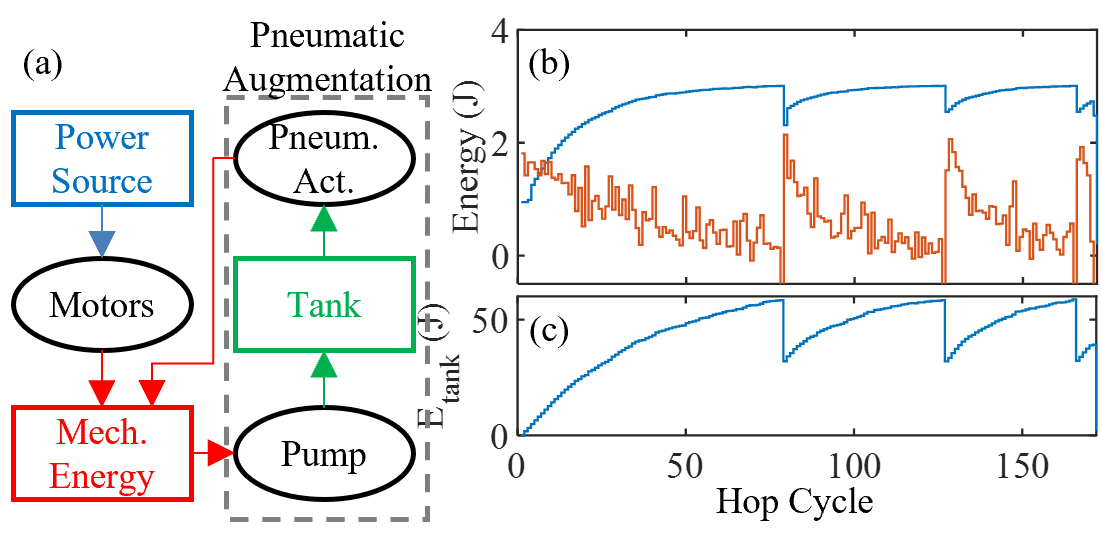}
    \caption{Energy analysis during hopping. (a) A block diagram representation of how energy is converted during a regular hop and augmented during an enhanced hop. Rectangles represent energy containers, ovals represent energy conversions. There is some loss associated with each arrow. (b) Estimations of the negative mechanical work done by the pump  $W_\text{pump}$ (blue) and the increase of pneumatic energy $\Delta E^k$ (red) in the tank over a number of hopping cycles. (c) The total energy stored in the tank  $E_\text{tank}$.}
    \label{fig:EnergyFig}
\end{figure}

Cyclic locomotion behaviors require cyclic power input and produce energy conversions. Fig. \ref{fig:EnergyFig} (a) displays how energy flows from the power source, converts into mechanical energy (M.E.) from the motors, accumulates to the pneumatic energy (P.E.) from the M.E. during regular hopping via the pump, and then returns to M.E. from P.E. through the pneumatic actuator for realizing enhanced hopping.  


The conversion from M.E. to P.E. can be estimated from the pump resistive force model and the pressure in the tank. Fig. \ref{fig:EnergyFig} (b) demonstrates the amount of negative work done by the pump at each hopping cycle and the increase of the energy stored in the tank in the form of air pressure. The negative work done by the pump to the robot during cycle $k$ is calculated as 
$W_\text{pump}=\int{F_\text{pump}(L)dL}$ and the energy added to the tank during cycle $k$ is calculated as 
$\Delta E^k_\text{tank} = V_\text{tank}(P^k_\text{tank} - P^{k-1}_\text{tank})$ \cite{poling2004properties}.
As the tank pressure increases, more mechanical work is done in the form of compressing air to bring the pump pressure up to the level of the tank pressure during descending; the compressed air in the pump is then dissipated during ascending. After enough periodic hops, the pump no longer can convert the negative work to P.E. as the robot is unable to compress the leg length enough due to the back pressure from the tank at the check valve. This indicates that there is a maximum amount of P.E. the robot can store. Fig.  \ref{fig:EnergyFig} (c) validates this by plotting the accumulation of energy in the tank in the form of pressure. 


As the tank reaches the desired pressure, the stored energy is released and converted to the M.E. of the robot through the pneumatic actuator, showing as instantaneous decreases in Fig. \ref{fig:EnergyFig} (c) for enhanced hops. 
The work done by the pneumatic actuator can be estimated by $W_\text{pa}=\int F_\text{pa}(L,t)dL$ assuming full leg extension from the fully compressed length. 
Yet, in reality, only a portion of P.E. is converted to M.E. through $W_\text{pa}$ due to delays in the solenoid valve in opening and closing. We use the experiments in Fig. \ref{fig:TrajectoriesFig} and estimate that the kinetic energy (K.E.) at lift-off (where the mechanical potential energy is the same) for the enhanced-free and enhanced hopping, showing an increase of K.E. from $4.66J$ to $18.31J$. All this validates that the pneumatic augmentation is capable of harvesting the M.E. from the periodic hops, and then return it back to the M.E. of the robot to perform high-power-output jumping behaviors.